%% file: main.tex
\documentclass{article} 
\usepackage{iclr2020_conference,times}

\input{math_commands.tex}

\usepackage{graphicx}

\usepackage{hyperref}
\usepackage{url}
\usepackage[caption=false]{subfig}

\title{Low resource language dataset creation, curation and classification: Setswana and Sepedi - Extended Abstract}

\author{Vukosi Marivate$^{1,2}$, 
      Tshephisho Sefara$^{2}$, 
      Vongani Chabalala$^{3}$, 
      Keamogetswe Makhaya$^{4}$,\\
      \textbf{Tumisho Mokgonyane$^{5}$, 
      Rethabile Mokoena$^{6}$, 
      Abiodun Modupe$^{7,1}$} \\
University of Pretoria, South Africa$^{1}$\\
CSIR, South Africa$^{2}$\\
University of Zululand, South Africa$^{3}$\\
University of Cape Town, South Africa$^{4}$\\
University of Limpopo, South Africa$^{5}$\\
North-West University, South Africa$^{6}$\\
University of the Witwatersrand, South Africa$^{7}$ \\
\texttt{vukosi.marivate@cs.up.ac.za} \\
\texttt{tsefara@csir.co.za}
}

\iclrfinalcopy 
\begin{document}
\maketitle

\begin{abstract}
The recent advances in Natural Language Processing have only been a boon for well represented languages, negating research in lesser known global languages. This is in part due to the availability of curated data and research resources. One of the current challenges concerning low-resourced languages are clear guidelines on the collection, curation and preparation of datasets for different use-cases. In this work, we take on the task of creating two datasets that are focused on news headlines (i.e short text) for Setswana and Sepedi and the creation of a news topic classification task from these datasets. In this study, we document our work, propose baselines for classification, and investigate an approach on data augmentation better suited to low-resourced languages in order to improve the performance of the classifiers.
\end{abstract}

\section{Introduction and Motivation}

The most pressing issues with regard to low-resource languages are lack of sufficient language resources, like features related to automation. In this study, we introduce an investigation of a low-resource language that provides automatic formulation and customisation of new capabilities from existing ones. While there are more than six thousand languages spoken globally, the availability of resources among each of those are extraordinarily unbalanced~\citep{nettle1998explaining}. For example, if we focus on language resources annotated on the public domain, as of November 2019, AG corpus released about $496,835$ news articles written in English language from more than $200$ sources ~\footnote{\url{http://groups.di.unipi.it/~gulli}}, Additionally, the Reuters News Dataset~\citep{lewis1997reuters} comprises out of roughly $10,788$ annotated texts from the Reuters financial newswire. Moreover, the New York Times Annotated Corpus holds over $1.8$ million articles~\citep{sandhaus2008new}. Lastly, Google Translate only supports around 100 languages \citep{johnson2017google}. A significant amount of knowledge exists for only on a small number of languages,
neglecting $17\%$ out of the world’s language categories labelled as low-resource, and there are currently no standard annotated tokens in low-resource languages \citep{strassel2016lorelei}. This in turn makes it challenging to develop various mechanisms and tolls used for Natural Language Processing (NLP).

In South Africa 
most of the news websites (private and public) are published in English, despite there being 11 official languages (including English). In this paper, we list the premium newspapers by circulation as per the first Quarter of 2019 (see Table~\ref{tab:circulation}).  Currently, there is a lack of information surrounding 8 of the 11 official South African languages, with the exception of English, Afrikaans and isiZulu which contain most of the reported datasets. In this work, we aim to provide a general framework for two of the 11 South African languages, to create an annotated linguistic resource for Setswana and Sepedi news headlines. In this study, we applied data sources of the news headlines from the South African Broadcast Corporation (SABC)~\footnote{\url{http://www.sabc.co.za/}}, their social media streams and a few acoustic news. Unfortunately, at the time of this study, we did not have any direct access to news reports, and hopefully this study can promote collaboration between the national broadcaster and NLP researchers. 

\begin{table}[]
\begin{center}

\caption{Top newspapers in South Africa with their languages}
\label{tab:circulation}
\begin{tabular}{|l|l|l|}
\hline
\textbf{Paper} & \textbf{Language} & \textbf{Circulation} \\ \hline
Sunday Times & English & 260132  \\ \hline
Soccer Laduma & English &  252041 \\ \hline
Daily Sun & English & 141187 \\ \hline
Rapport & Afrikaans & 113636 \\ \hline
Isolezwe & isiZulu  & 86342 \\ \hline
Sowetan & English  & 70120 \\ \hline
\end{tabular}
\end{center}
\end{table}

\section{News title data collection and annotation}
The news data is collected from SABC radio stations Thobela FM~\footnote{\url{https://www.facebook.com/thobelafmyaka/}} (Sepedi) and Motsweding FM~\footnote{\url{https://www.facebook.com/MotswedingFM/}} (Setswana). 
We collected 219 news headlines in Setswana and 491 news headlines in Sepedi. The datasets we used are relatively small and as such, we had to look at other ways to build vectorizers that can better generalise content, as the word token diversity would be very low in the sample. 

We categorised the news headlines into: \emph{Legal}, \emph{General News},\emph{Sports}, \emph{Other}, \emph{Politics}, \emph{Traffic News}, \emph{Community Activities}, \emph{Crime}, \emph{Business} and \emph{Foreign Affairs}. These categories, through definition by the authors, were best matched to fit headlines within both datasets. 
For this paper, we only explored single label categorisation for each article. It remains to be investigated in future work to explore multi-label case, which will assist the process by reducing noise in the analysis. To combat this, we created larger pre-trained vectorizers to have classifiers that generalise better. To achieve this, we collected a set of single language corpora (from Wikipedia, the bible, JW300 etc.)  for each language to create Bag of Words, TFIDF, Word2vec and FastText vectorizers. 

\section{News title classification}

We performed 5 fold cross validation experiments on classifying the news articles titles. Furthermore, we investigated the use of Logistic Regression (LR), Support Vector Classification (SVC), XGBoost classifiers and a Multi-layer Perceptron (MLP) on the task. We further investigated the use of word2vec (contextual) based augmentation, which has been shown to be an effective approach that works well on short text analysis \citep{marivate2019improving}. The results are illustrated in Figure~\ref{fig:result_tf_tfidf} below.



\begin{figure*}[h]
\centering
    \subfloat[ Setswana\label{fig:tfsetswana}]{%
        \includegraphics[width=0.45\linewidth]{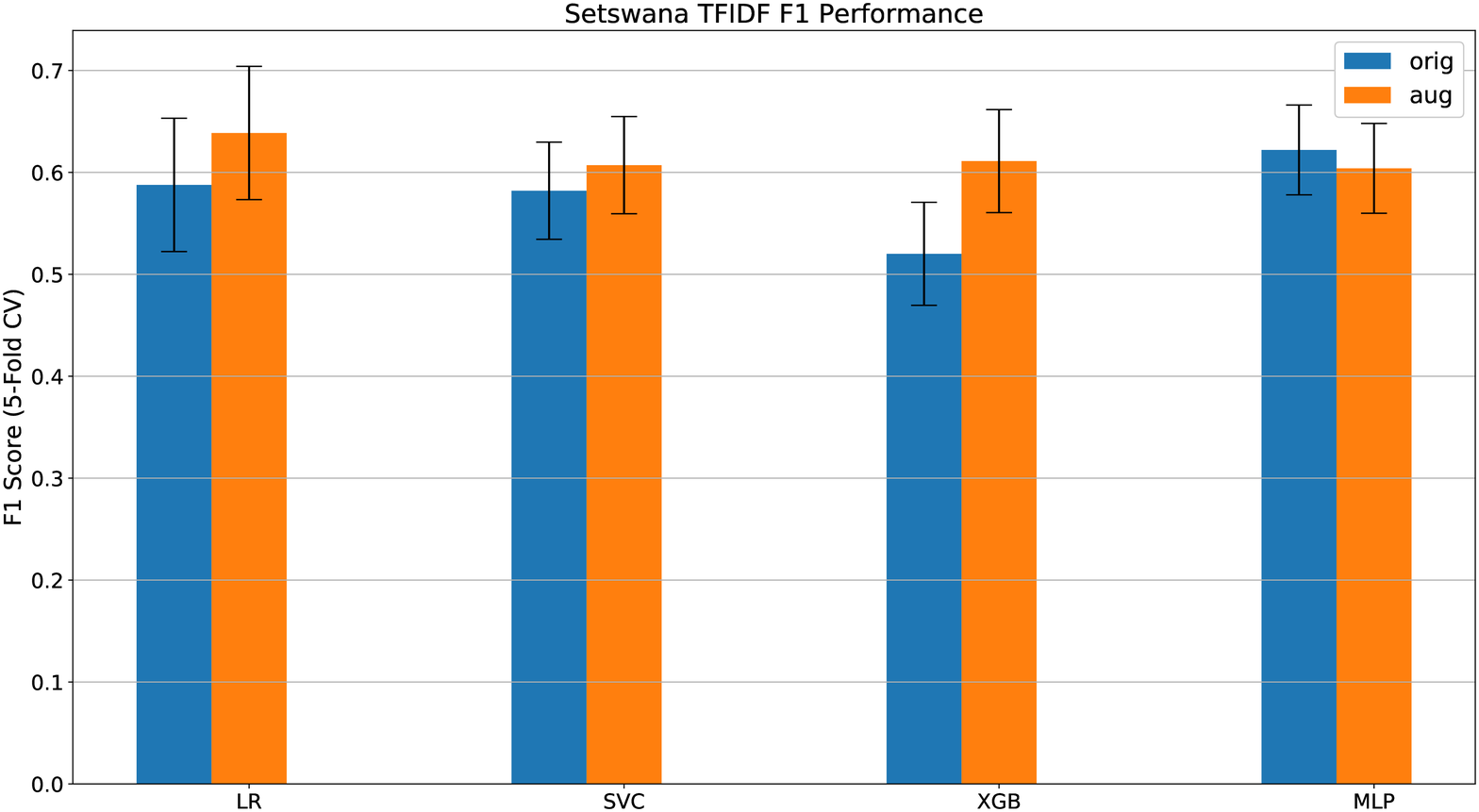}
    }
    \subfloat[ Sepedi\label{fig:tfsepedi}]{%
        \includegraphics[width=0.45\textwidth]{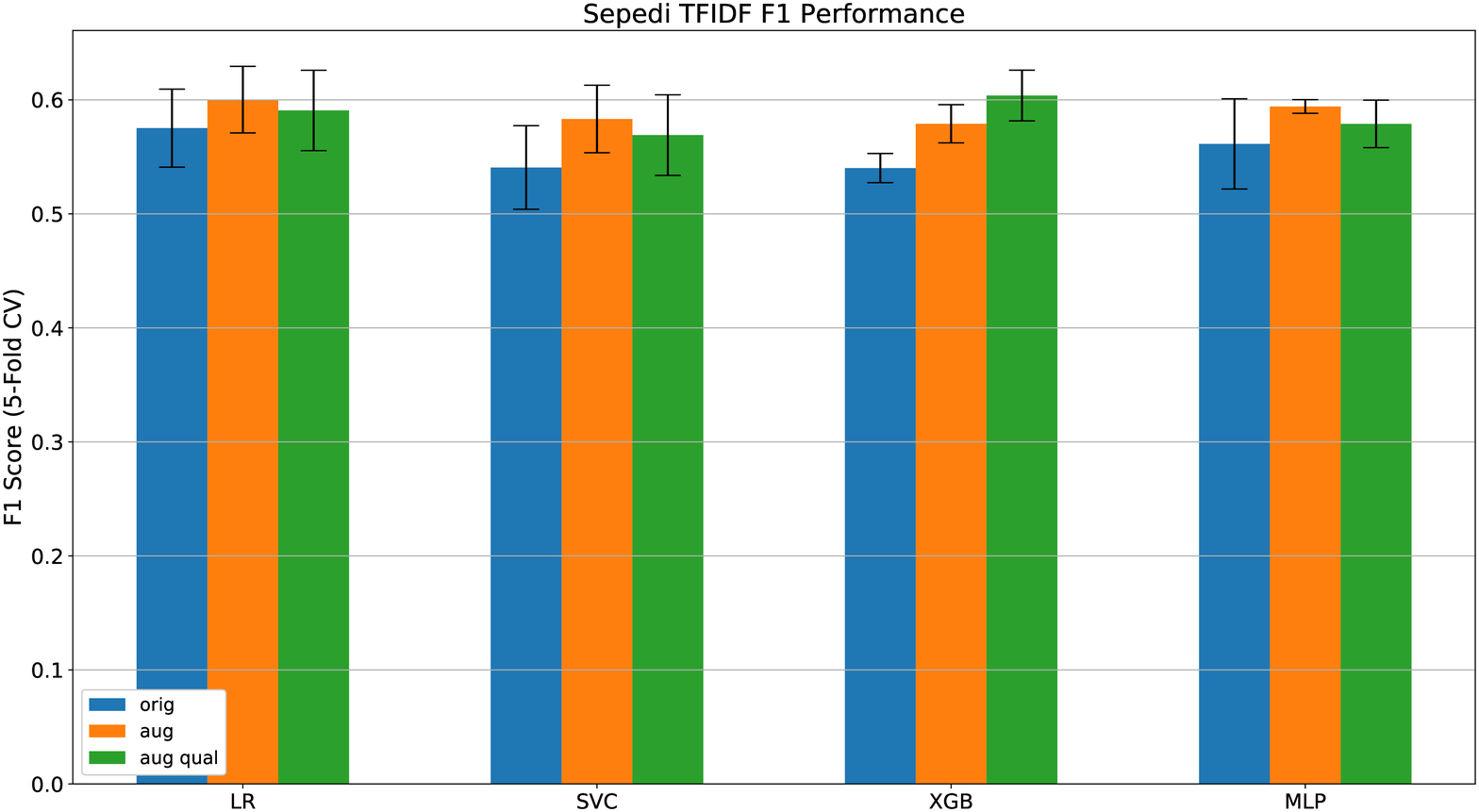}
    } 
    \caption{Classification model performance for news title categorisation in Setswana and Sepedi.}\label{fig:result_tf_tfidf}
\end{figure*}

\section{Future Work}

On the data side, we are working on the release of both pre-trained embeddings and to release the news headline data for all 9 South African languages, publicly. Lastly, we are working on improving our classification models and investigating deep learning approaches in our methodology.

\bibliographystyle{iclr2020_conference}
\bibliography{references}
\end{document}

%% file: math_commands.tex
\usepackage{amsmath,amsfonts,bm}

\def\eqref#1{equation~\ref{#1}}

\def\1{\bm{1}}

\DeclareMathAlphabet{\mathsfit}{\encodingdefault}{\sfdefault}{m}{sl}
\SetMathAlphabet{\mathsfit}{bold}{\encodingdefault}{\sfdefault}{bx}{n}

